\documentclass[twoside,twocolumn,10pt]{article}
 
\usepackage{wscg}           
\RequirePackage{ifpdf}
\ifpdf
 \RequirePackage[pdftex]{graphicx}
 \RequirePackage[pdftex]{color}
\else
 \RequirePackage[dvips,draft]{graphicx}
 \RequirePackage[dvips]{color}
\fi
 
\usepackage{nopageno}       
 
\usepackage{amsmath,amssymb}
\usepackage{booktabs}
\usepackage{algorithmic}
\usepackage{algorithm}

\usepackage{fancyhdr}

\usepackage{caption}
\renewcommand{\topfraction}{0.95}

\renewcommand{\floatpagefraction}{0.75}

\usepackage[pscoord]{eso-pic}
\usepackage[color=black,opacity=1,angle=0,scale=1]{background}
\newcommand{\placetextbox}[3]{
  \setbox0=\hbox{#3}
  \AddToShipoutPictureFG*{
    \put(\LenToUnit{#1\paperwidth},\LenToUnit{#2\paperheight}){\vtop{{\null}\makebox[0pt][c]{#3}}}%
  }%
}%
\backgroundsetup{
  contents={%
  \placetextbox{0.5}{.95}{Accepted WSCG 2026}
  \placetextbox{0.5}{0.085}{\thepage}%
  }
}

\title{SC-MFJ: A Simple Haptic Quality Metric\\for Medical Image Segmentation}
 
\author{
\parbox{0.28\textwidth}{\centering
Souraj Adhikary\\[1mm]
Jade University of Applied Sciences\\
Friedrich-Paffrath-Str.\ 101\\
26389 Wilhelmshaven, Germany\\[1mm]
souraj.adhikary@student.jade-hs.de
}
\hspace{0.03\textwidth}
\parbox{0.28\textwidth}{\centering
Negar Chabi\\[1mm]
Jade University of Applied Sciences\\
Friedrich-Paffrath-Str.\ 101\\
26389 Wilhelmshaven, Germany\\[1mm]
negar.chabi@jade-hs.de
}
\hspace{0.03\textwidth}
\parbox{0.28\textwidth}{\centering
Andre Mastmeyer\\[1mm]
Jade University of Applied Sciences\\
Friedrich-Paffrath-Str.\ 101\\
26389 Wilhelmshaven, Germany\\[1mm]
andre.mastmeyer@jade-hs.de
}
}
 
\usepackage{url}
\makeatletter
\def\Uslash{\mathbin{\mathchar`\/}\@ifnextchar{/}{\kern-.15em}{}}
\g@addto@macro\UrlSpecials{\do \/ {\Uslash}}
\def\Ucolon{\mathbin{\mathchar`:}\@ifnextchar{/}{\kern-.1em}{}}
\g@addto@macro\UrlSpecials{\do : {\Ucolon}}
\makeatother

\begin{document}
 
\twocolumn[{\csname @twocolumnfalse\endcsname
 
\maketitle  

\begin{abstract}
\noindent
Standard segmentation metrics such as Dice and Hausdorff distance
measure geometric overlap but say nothing about whether a segmented
surface is suitable for haptic rendering in surgical simulation.
We propose SC-MFJ (Surface-Constrained Mean Force Jerk), a simple,
inexpensive metric that samples a segmented organ surface with many
short virtual stylus walks and measures how jerky the resulting
contact forces are.
The metric is computed from existing segmentation outputs and uses
roughly one minute of CPU time per case.
We evaluate three pancreas CT segmentation approaches---binary nnU-Net
output, Gaussian-smoothed output, and learned signed distance function
(SDF) regression---across 80 cases in five-fold cross-validation.
SC-MFJ reveals a $147\times$ gap in haptic quality between the raw
binary baseline and simple Gaussian post-processing, a difference
entirely invisible to Dice and HD95.
It also shows that learned SDF regression, despite requiring full model
retraining, produces more variable haptic quality than Gaussian smoothing,
with a case-level standard deviation of $168$~N/s$^2$ compared with
$22$~N/s$^2$ for Gaussian.
A second evaluation on the LiTS liver dataset (131 cases) confirms
the generality of these findings: the binary-to-Gaussian gap widens
to $189\times$, and Gaussian smoothing again produces consistently
low force jerk across all folds.
Our results suggest that for haptic simulation applications,
a one-line post-processing step may be sufficient, and that a cheap
metric like SC-MFJ can flag problems that geometric metrics miss.
\end{abstract}
 
\subsection*{Keywords}
Haptic rendering, segmentation quality, force jerk, signed distance function, surgical simulation.
 
\vspace*{1.0\baselineskip}
}]

 
\section{Introduction}
\label{sec:intro}
 
\copyrightspace
 
Specialized endoscopic techniques such as Endoscopic Ultrasound (EUS) allow clinicians to visualize the pancreas and surrounding structures with high-resolution imaging.
These procedures require precise instrument control in the vicinity of critical vascular structures, and complication rates correlate with operator experience.
Visuo-haptic VR simulators offer a way to build procedural competence without patient risk, but their fidelity depends on the quality of the underlying organ model: voxel-level staircase artifacts from standard segmentation pipelines produce force discontinuities that degrade the training experience and may teach incorrect tissue interaction patterns.
The gap between what segmentation pipelines produce and what haptic renderers need has received little systematic attention.
 
In surgical simulation, CT-derived organ models are rendered at
haptic update rates of 1~kHz~\cite{Rizzi2012}.
The pipeline from scan to simulator typically runs through three
stages: image segmentation, surface extraction (e.g.\ Marching
Cubes~\cite{Lorensen1987}), and penalty-based force computation at the surface.
Each stage is evaluated independently.
Segmentation researchers report Dice coefficients and Hausdorff
distances~\cite{Isensee2021}.
Haptic engineers run user studies and measure device
stability~\cite{Colgate1994}.
The two communities rarely compare notes.
 
Maier-Hein et al.~\cite{MaierHein2024} recently documented this
disconnect in their \emph{Metrics Reloaded} framework, showing that
chosen validation metrics in biomedical image analysis often fail to
reflect the actual domain interest.
They call for application-specific metrics---but do not propose one
for haptic rendering.
 
We offer a small step in that direction. SC-MFJ (Surface-Constrained Mean Force Jerk) is a simple metric
that evaluates a segmented surface by simulating short virtual stylus
walks over it and measuring the resulting force jerk---the second time
derivative of force, capturing how abruptly contact forces change
direction.
The walks are not meant to reproduce one specific surgical maneuver;
they are a way to sample whether the surface itself would generate
smooth or discontinuous force feedback.
This builds on the minimum-jerk smoothness criterion
from motor control~\cite{Flash1985}, subsequently adopted
as a widely used metric in surgical skill evaluation~\cite{Estrada2014}.
The metric is cheap to compute (about one minute of CPU time per
case), is computed directly from segmentation outputs, and can be
applied to any output that produces an isosurface.
 
The contributions of this paper are:
\begin{enumerate}
  \item SC-MFJ, a physics-grounded metric that connects segmentation
        surface quality to haptic rendering suitability, together with
        a characterization of its formal properties.
  \item An empirical comparison on the NIH Pancreas CT
        dataset~\cite{Roth2015} and the LiTS liver tumor
        segmentation benchmark~\cite{Bilic2023} showing that SC-MFJ reveals quality
        differences invisible to Dice and HD95, with consistent
        results across two organs.
  \item A Gaussian $\sigma$ sweep demonstrating that the
        Dice--SC-MFJ trade-off follows a Pareto curve with a
        practical operating point at $\sigma = 1.0$, and a correlation
        analysis confirming that SC-MFJ captures force-rendering
        smoothness not measured by Dice.
\end{enumerate}
 
We do not claim that SC-MFJ replaces geometric metrics.
It measures a different thing---force rendering smoothness rather than spatial
overlap---and is most useful when the downstream application involves
physical simulation.

\section{Related Work}
\label{sec:related}
 
\textbf{Segmentation metrics.}
The Dice coefficient and Hausdorff distance (HD95) remain the dominant
evaluation metrics in medical image segmentation~\cite{MaierHein2024}.
Task-motivated alternatives exist: Nikolov et al.~\cite{Nikolov2021}
proposed surface Dice for contour-editing workflows, and
Shit et al.~\cite{Shit2021} introduced clDice for tubular
connectivity.
These metrics are tailored to specific clinical tasks but remain
geometric---none evaluates physical forces.
 
\textbf{SDF-based segmentation.}
Xue et al.~\cite{Xue2020} proposed predicting signed distance maps
(SDMs) to produce smoother organ boundaries.
They report improvements via Dice and average surface distance but
do not evaluate force-based smoothness.
In a surgical context, Nysj\"{o} et al.~\cite{Nysjo2017} used
anti-aliased signed distance fields computed from CT bone
segmentations for generating surgical guides and plates, but
likewise did not evaluate the resulting surfaces in terms of
haptic force quality.
Ma et al.~\cite{Ma2020} benchmarked distance transform methods and
documented convergence failures, noting that distance regression
``may not converge'' without coupling to Dice loss.
In a related but distinct setting, Yang et al.~\cite{Yang2023}
show that neural SDF optimization is unstable in the continuum
limit for 3D shape reconstruction.
Their analysis does not address voxel-based medical SDF regression,
but the underlying optimization challenge---fitting a continuous
distance field with a neural network---is shared, and the theoretical
instability is consistent with the empirical convergence failures
observed in the medical domain~\cite{Ma2020}.
 
\textbf{Haptic rendering and jerk.}
In surgical simulation, jerk---the third temporal derivative of
position---is a widely used measure of movement
smoothness~\cite{Flash1985}.
Haptic rendering in surgical contexts requires stable force feedback
at update rates of 1~kHz or higher; the foundations and algorithms
for such rendering are surveyed by Lin and Otaduy~\cite{Lin2008}.
It is routinely used to assess \emph{operator} skill in
surgical training~\cite{Estrada2014,Shayan2023}.
However, it has not been applied to evaluate the \emph{input
surface} that the operator interacts with.
Chan et al.~\cite{Chan2016} note that anatomic segmentation is
a key component of their temporal bone simulator pipeline but
evaluate simulator fidelity primarily through visual comparison
with intraoperative video, not through quantitative force metrics.
Rizzi et al.~\cite{Rizzi2012} defined a force anomaly coefficient
based on force derivatives to evaluate haptic rendering
\emph{algorithms} on isosurfaces extracted from volumetric data.
Mastmeyer et al.~\cite{Mastmeyer2013} proposed a ray-casting-based
evaluation framework that systematically compares needle insertion
force feedback algorithms along thousands of automatically generated
paths, establishing the need for quantitative force output evaluation
in puncture simulation.
Fortmeier et al.~\cite{Fortmeier2013} developed optimized image-based
soft tissue deformation algorithms for real-time visualization of
haptic needle insertion, addressing the rendering requirements of
visuo-haptic surgical simulators.
Mastmeyer et al.~\cite{Mastmeyer2012} applied anisotropic diffusion
filtering to CT volumes prior to direct haptic volume rendering,
demonstrating that spatial smoothing can reduce force artifacts in
puncture simulation; however, the improvement was assessed
qualitatively rather than with a force-derivative metric.
SC-MFJ applies a related principle---force-derivative analysis---to
evaluate the input segmentation surface itself.
We compute the second temporal derivative of force ($d^2\mathbf{F}/dt^2$),
which we use operationally as \emph{force jerk}: it measures abrupt
curvature and oscillation in the force trajectory.
This is not kinematic jerk, which is the third derivative of position;
the term reflects the analogous smoothness role of the quantity for
force feedback rather than its derivative order.
We evaluate this quantity along surface-constrained random walks rather than
operator-driven trajectories.

\section{Method}
\label{sec:method}
 
SC-MFJ measures how jerky the contact forces would be if a virtual
haptic stylus were dragged along a segmented organ surface.
The idea is deliberately simple: sample short paths on the surface,
record the contact force along each path, and compute how sharply the
force changes from one step to the next.
Throughout the method section, bold symbols denote vectors.
Thus $\mathbf{p}_i$, $\hat{\mathbf{n}}_i$, $\mathbf{F}_i$, and
$\mathbf{J}_i$ are vector quantities, while SC-MFJ is scalar because it
averages the magnitudes $\|\mathbf{J}_i\|$.
 
\subsection{Overview}
 
Given a 3D scalar field $\phi$ (a signed distance field, a smoothed
distance transform, or a raw binary distance transform) and a level
value $\ell$ (typically $0$), the zero-level set
$\{\mathbf{x} : \phi(\mathbf{x}) = \ell\}$ defines the organ surface.
SC-MFJ then runs $N$ short random walks on this surface.
A walk starts at a randomly selected surface point, moves in a random
tangent direction, and is projected back onto the surface after every
step.
This gives a procedure-agnostic sample of many local surface regions
rather than one hand-picked instrument path.
For each sampled point, SC-MFJ computes a contact force from the local
surface normal and then averages the force jerk magnitude across all
valid steps of all walks.
 
\subsection{Surface Walk and Force Model}
 
At each point $\mathbf{p}_i$ along a walk, we estimate the surface
normal from the gradient of the scalar field.
For a clean surface, neighboring normals change gradually.
For a voxel-staircase surface, neighboring normals can flip abruptly.
This normal field is therefore the part of the segmentation that most
directly affects the force signal:
\begin{equation}
  \hat{\mathbf{n}}_i = \frac{\nabla\phi(\mathbf{p}_i)}
                             {\|\nabla\phi(\mathbf{p}_i)\|}
  \label{eq:normal}
\end{equation}
The contact force is modeled with a minimal spring penalty.
We do not simulate tissue deformation or a full haptic device.
Instead, a fixed penetration depth $\delta$ and stiffness $k$ convert
the local surface normal into a restoring force:
\begin{equation}
  \mathbf{F}_i = -k \cdot \delta \cdot \hat{\mathbf{n}}_i
  \label{eq:force}
\end{equation}
Because $k$ and $\delta$ are constants shared across all methods,
the only quantity that changes along a walk is the local normal.
SC-MFJ therefore measures how smoothly the surface normal changes
along the sampled path.
If the normals vary slowly, forces change slowly, and jerk is low.
If staircase artifacts make the normals oscillate, the force direction
also oscillates, and jerk is high.
 
\subsection{Jerk Computation}
 
Force jerk is computed from the force time series along each walk.
We use the second finite difference because it penalizes rapid changes
in the direction or curvature of the force signal, not just the force
magnitude itself.
The simulated haptic update rate is 1~kHz (time step
$\Delta t = 0.001$~s):
\begin{equation}
  \mathbf{J}_i = \frac{\mathbf{F}_{i+1} - 2\mathbf{F}_i + \mathbf{F}_{i-1}}
                      {\Delta t^2}
  \label{eq:jerk}
\end{equation}
The SC-MFJ score for one case is the mean of $\|\mathbf{J}_i\|$
across all valid steps of all $N$ walks.
Large isolated spikes and repeated small oscillations both increase
this average, which is useful for detecting surfaces that would feel
rough in force feedback even when their Dice score is high.
 
\subsection{Surface-Constrained Design}
 
A key design choice is that the stylus never leaves the surface.
At every step, we first move a small distance in the current tangent
direction, producing a candidate point.
That candidate is then \emph{projected back} onto the isosurface.
We first attempt a Newton correction along the gradient direction;
if this fails, we snap to the nearest mesh vertex via a KD-tree.
This prevents the walk from drifting into the volume or away from the
organ, where the force would no longer represent surface contact.

The random walks are used as a sampling rule, not as a model of a
specific operation.
Each walk starts from a randomly selected mesh vertex, so different
regions of the organ surface are sampled instead of a single chosen
curve.
All methods---binary, Gaussian, SDF---are evaluated under the same
protocol: identical random seed, identical number of walks, identical
trajectory length, and identical step size.
Because the meshes differ between methods, start points are drawn from
each method's own mesh; what is kept fixed is the sampling procedure.
Differences in SC-MFJ therefore reflect how smooth the evaluated
surface is under the same test, not whether one method received an
easier hand-picked trajectory.
 
Note that spatial smoothing of a segmentation volume does not
mathematically guarantee low force jerk on the extracted isosurface.
Aggressive smoothing can introduce topological changes that create
new force discontinuities.
The connection between Gaussian smoothing and reduced SC-MFJ is an
empirical finding, not a circular one.

\clearpage
\subsection{Pseudocode}
 
Algorithm~\ref{alg:scmfj} summarizes the procedure.
 
\begin{algorithm}[ht!]
\caption{SC-MFJ computation with random surface walks}
\label{alg:scmfj}
\begin{algorithmic}[1]
\REQUIRE scalar field $\phi$, spacing, level $\ell$, $N$ trajectories
\STATE Extract mesh at level $\ell$ via Marching Cubes~\cite{Lorensen1987}
\STATE Build KD-tree over mesh vertices
\STATE Precompute gradient field $\nabla\phi$
\FOR{$n = 1$ to $N$}
  \STATE Pick random mesh vertex as start $\mathbf{p}_0$
  \STATE Sample random tangent direction $\mathbf{u}$
  \FOR{each step $i$}
    \STATE $\hat{\mathbf{n}}_i \leftarrow \nabla\phi(\mathbf{p}_i) / \|\nabla\phi(\mathbf{p}_i)\|$
    \STATE $\mathbf{F}_i \leftarrow -k \cdot \delta \cdot \hat{\mathbf{n}}_i$
    \STATE $\mathbf{p}_{\text{cand}} \leftarrow \mathbf{p}_i + \Delta x \cdot \mathbf{u}$
    \STATE Reproject $\mathbf{p}_{\text{cand}}$ onto surface (Newton or KD-tree)
    \STATE Update tangent direction $\mathbf{u}$ (remove normal component)
  \ENDFOR
  \STATE Compute jerk: $\mathbf{J}_i = (\mathbf{F}_{i+1} - 2\mathbf{F}_i + \mathbf{F}_{i-1}) / \Delta t^2$
\ENDFOR
\RETURN mean of $\|\mathbf{J}_i\|$ over all valid steps and walks
\end{algorithmic}
\end{algorithm}

\subsection{Formal Properties}
\label{sec:formal}

We summarize the formal properties of SC-MFJ that follow from its
definition (Eqs.~\ref{eq:force}--\ref{eq:jerk}).

\textbf{Non-negativity.}
SC-MFJ is defined as the mean of force jerk magnitudes
$\|\mathbf{J}_i\|$, each of which is a norm and therefore
non-negative.
Hence SC-MFJ~$\geq 0$ for any surface.

\textbf{Zero condition.}
On a planar surface, the surface normal $\hat{\mathbf{n}}$ is
constant along any walk.
Because $k$ and $\delta$ are constants, the force
$\mathbf{F}_i = -k \cdot \delta \cdot \hat{\mathbf{n}}$ is constant
at every step, yielding
$\mathbf{J}_i = \mathbf{0}$ for all $i$ and thus SC-MFJ~$= 0$.
Planarity is therefore a sufficient condition for a zero score.
More generally, SC-MFJ is zero whenever the sampled force sequence
has zero second finite difference, so the metric should not be
interpreted as a formal test of planarity.

\textbf{Undefined on empty segmentation.}
If the segmentation mask is empty, no isosurface exists and no mesh
vertices can be sampled.
SC-MFJ is undefined in this case and should not be reported.

\textbf{Behaviour under Gaussian smoothing.}
Gaussian convolution of the scalar field is expected to reduce
high-frequency normal variation and therefore force jerk.
This is not a formal monotonicity guarantee: aggressive smoothing can
shift the isosurface or introduce topological changes that create new
force discontinuities.
We therefore treat the effect of $\sigma$ as an empirical question,
evaluated in the Gaussian sweep in Section~\ref{sec:sigma}.

\textbf{Linear scaling with contact parameters.}
Because force is $\mathbf{F}_i = -k \cdot \delta \cdot \hat{\mathbf{n}}_i$
and jerk is the second finite difference of force, SC-MFJ scales
linearly with the product $k \cdot \delta$ for a fixed sampled path.
Changing $k$ or $\delta$ therefore rescales the absolute value of the
metric but does not change the ordering induced by the surface normals.

\textbf{Computational complexity.}
For $N$ trajectories of $L$ steps each on a mesh with $V$ vertices,
SC-MFJ requires $O(N \cdot L)$ force evaluations plus
$O(V \log V)$ for KD-tree construction.
In practice, the total runtime is approximately one minute per case
on a single CPU core.

\section{Experimental Setup}
\label{sec:setup}
 
\subsection{Dataset}
We use the NIH Pancreas CT dataset~\cite{Roth2015}, comprising
82~abdominal CT volumes with manual pancreas annotations.
We perform five-fold cross-validation over 80 cases
(16 per fold).%
\footnote{Two of the 82 cases were excluded due to data corruption
that prevented reliable prediction across all three methods.
All methods are evaluated on the same 80-case set.}

To test whether SC-MFJ generalises beyond a single organ, we
additionally evaluate on the Liver Tumor Segmentation Benchmark
(LiTS)~\cite{Bilic2023}, comprising 131 abdominal CT volumes with
liver annotations.
The segmentation model is used only as a consistent surface generator
for SC-MFJ evaluation, not as the contribution of this work.
We use the same 3D full-resolution setup over all 131 LiTS cases.
Only the binary baseline and Gaussian $\sigma = 1.0$ conditions are
evaluated on LiTS; SDF regression was not included because the
training instability analysis is already reported on the primary
pancreas dataset.
 
\subsection{Segmentation Methods}
We use nnU-Net~\cite{Isensee2021} as a common segmentation backbone
so that SC-MFJ compares surface representations rather than different
network architectures.
We evaluate three conditions:
 
\begin{itemize}
  \item \textbf{Binary baseline}: Standard nnU-Net classification
        output (hard 0/1 voxel labels), converted to a truncated
        signed distance field via \texttt{SignedMaurerDistanceMap}
        for SC-MFJ evaluation.
 
  \item \textbf{Gaussian $\sigma$=1.0}: The binary baseline's
        distance field is smoothed with a Gaussian kernel
        ($\sigma = 1.0$ voxel on the nnU-Net resampled grid; the
        median resampled spacing is approximately
        $0.8 \times 0.8 \times 1.5$~mm).
        The smoothed zero-level set is then evaluated.
        This is a single line of post-processing code; no
        retraining is involved.
 
  \item \textbf{SDF regression}: The nnU-Net backbone is modified
        to produce a single-channel output mapped to $[-1, 1]$ via
        $\tanh$ activation and trained to regress a continuous signed
        distance field against precomputed truncated SDF targets
        derived from the ground-truth labels, following the idea of
        Xue et al.~\cite{Xue2020}.
        Training uses a composite regression objective that
        couples field-value accuracy with an overlap term for
        geometric convergence.
        The native SDF output is post-processed with
        largest-connected-component extraction.
        SDF regression is a natural candidate when smoother organ
        boundaries are desired for haptic rendering; however, it
        requires full model retraining and is known to suffer from
        convergence instability~\cite{Ma2020}.
\end{itemize}

Because all segmentation variants emerge from the same nnU-Net
framework as the baseline, the comparison is controlled: all three
methods share the same backbone architecture and differ only in output
representation and post-processing.

\subsection{SDF Training Protocol}
Folds~0, 1, and 4 converged
with the initial random seed. Folds~2 and 3 required seed
reselection; in each case the first converging seed was used, with
model selection based solely on validation Dice. No SC-MFJ scores
were consulted during training or checkpoint selection. SC-MFJ
evaluation was performed post-hoc on frozen models. SDF training
instability is a well-documented phenomenon in the
literature~\cite{Ma2020}.
 
\subsection{SC-MFJ Parameters}
We use $N = 50$ trajectories per case, trajectory length
$20$~mm, stylus speed $50$~mm/s (yielding a 1~kHz update rate
with $\Delta x = 0.05$~mm steps), stiffness $k = 1.0$~N/mm,
penetration depth $\delta = 0.5$~mm, and a fixed random seed of~42.
These parameters define the measurement protocol: how much of the
surface is sampled, how densely the path is sampled, and how the local
normal is converted into force.
Because $k$ and $\delta$ are held constant across all methods,
absolute SC-MFJ values depend on this parameter choice, but relative
comparisons are unaffected.
The contact-parameter and trajectory-count analyses in
Sections~\ref{sec:sensitivity} and~\ref{sec:convergence} check that the
main conclusions are not driven by these choices.
All experiments were run on a single CPU core (AMD EPYC 7713) under AlmaLinux~8.10 using Python~3.11, SimpleITK~2.5.3, SciPy~1.17.0, and NumPy~2.4.2.
 
\subsection{Geometric Metrics}
Dice coefficient and 95th-percentile Hausdorff distance (HD95) are
computed on the binary segmentation masks for all three methods.
 
\begin{figure*}[t]
  \centering
  \includegraphics[width=\textwidth]{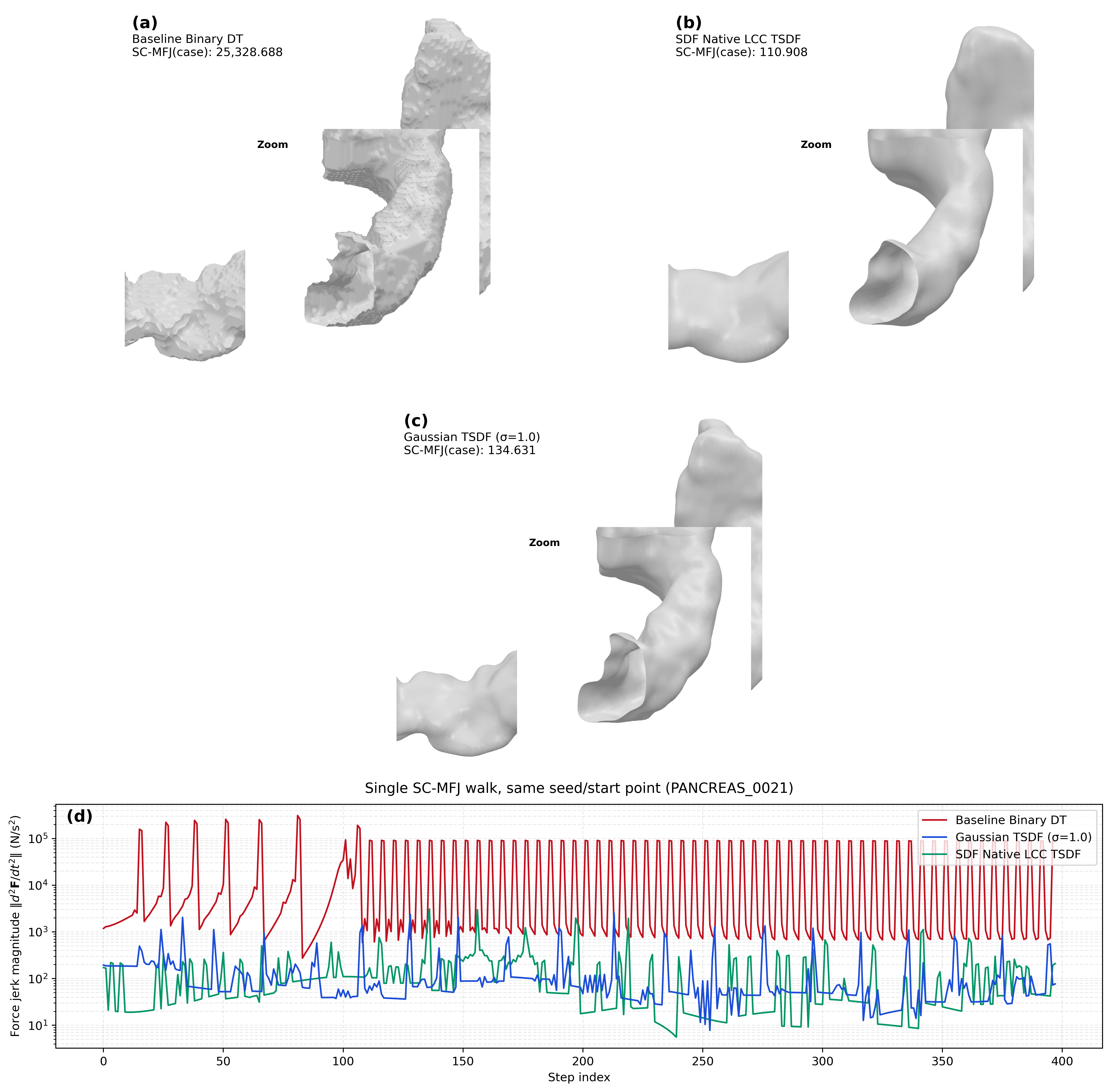}
  \caption{Representative case PANCREAS\_0021, selected because
  Gaussian and SDF achieve comparable Dice, isolating surface
  quality. (a)~Binary baseline: SC-MFJ = 25,329.
  (b)~SDF Native LCC: SC-MFJ = 111.
  (c)~Gaussian $\sigma$=1.0: SC-MFJ = 135, highlighted in the
  bottom row to emphasize the main result.
  Zoom insets highlight staircase artifacts on the binary surface.
  (d)~Force jerk magnitude along one surface walk (log scale).
  The binary baseline (red) oscillates two orders of magnitude
  higher than Gaussian (blue) and SDF (green).
  In this particular case, SDF slightly outperforms
  Gaussian---the overall advantage of Gaussian lies in its
  cross-fold consistency, not in every individual case.
  Each surface walk contains 400 steps because the path length is
  20~mm and the step size is $\Delta x = 0.05$~mm; SC-MFJ averages
  50 such walks per case.}
  \label{fig:example}
\end{figure*}

\section{Results}
\label{sec:results}
 
\subsection{Main Comparison}
 
Table~\ref{tab:main} shows the aggregate results across all 80
validation cases.
In terms of Dice and HD95, the three methods are broadly similar:
Dice ranges from 0.838 to 0.864, and HD95 from 4.2 to 5.0~mm.
A researcher using only these metrics would conclude that the methods
differ only modestly geometrically, even though their haptic rendering
behaviour differs substantially.
 
SC-MFJ tells a different story.
The binary baseline produces an SC-MFJ of $23{,}374$~N/s$^2$,
reflecting severe force oscillations caused by staircase artifacts
on the voxel grid.
Gaussian smoothing reduces this to $159$~N/s$^2$---a
$147\times$ improvement---while SDF regression achieves
$197$~N/s$^2$.
This gap is invisible to Dice and HD95.
A Wilcoxon signed-rank test on 80 paired cases found no significant
difference between Gaussian and SDF SC-MFJ
($W = 1319$, $p = 0.149$; Gaussian lower in 44 of 80 cases).
The two methods thus produce comparable mean haptic quality; the
critical distinction is consistency, examined in the next section.
 
\begin{table}[htb]
\centering
\caption{Aggregate results across 80 validation cases
(five-fold cross-validation). Higher Dice is better; lower HD95 and
SC-MFJ are better. For SC-MFJ, lower standard deviation indicates
greater case-level consistency; bold marks the most consistent smooth
method.}
\label{tab:main}
\vspace{2mm}
\small
\resizebox{\columnwidth}{!}{%
\begin{tabular}{@{}lccc@{}}
\toprule
Method & Dice & HD95 (mm) & SC-MFJ (N/s$^2$) \\
\midrule
Binary baseline  & $0.864 \pm 0.038$ & $4.24 \pm 3.56$ & $23{,}374 \pm 3{,}153$ \\
Gaussian $\sigma$=1.0 & $0.838 \pm 0.039$ & $4.54 \pm 3.27$ & $159 \pm \textbf{22}$ \\
SDF regression   & $0.848 \pm 0.045$ & $4.96 \pm 3.48$ & $197 \pm 168$ \\
\bottomrule
\end{tabular}}
\end{table}
 
\subsection{Fold-Level Consistency}
 
Table~\ref{tab:folds} breaks down SC-MFJ by fold.
Gaussian smoothing produces remarkably stable results: SC-MFJ
ranges from 154 to 165 across folds (a spread of 11~N/s$^2$).
SDF regression ranges from 107 to 301 (a spread of
194~N/s$^2$)---nearly 18 times the variability.
 
This variance tracks with training instability.
SDF training is sensitive to initialization: two of five folds
required seed reselection to achieve geometric convergence at
all~\cite{Ma2020}.
Even the folds that converged geometrically show large differences
in surface quality.
A user of an SDF-based pipeline cannot predict from Dice alone
whether a given model will produce smooth or jerky force feedback.
Gaussian smoothing, by contrast, requires no model retraining and
produces consistent haptic quality across every fold.
 
The key observation is the dramatically lower standard deviation
of Gaussian smoothing: $\pm 22$~N/s$^2$ in Table~\ref{tab:main}
and a range of only 11~N/s$^2$ across folds in
Table~\ref{tab:folds}, compared to $\pm 168$~N/s$^2$ and a range
of 194~N/s$^2$ for SDF regression.
This consistency is critical for haptic applications, where
unpredictable force behaviour across training runs is unacceptable.
 
\begin{table}[htb]
\centering
\caption{SC-MFJ by fold for the two smooth methods.
Lower range indicates greater fold-level consistency; bold marks the
more consistent method.}
\label{tab:folds}
\vspace{2mm}
\small
\begin{tabular}{@{}ccc@{}}
\toprule
Fold & Gaussian SC-MFJ & SDF SC-MFJ \\
\midrule
0 & 159.96 & 163.19 \\
1 & 154.12 & 213.92 \\
2 & 160.59 & 301.05 \\
3 & 165.00 & 250.66 \\
4 & 156.23 & 107.28 \\
\midrule
Range & \textbf{11} & 194 \\
\bottomrule
\end{tabular}
\end{table}

\begin{figure*}[!t]
  \centering
  \includegraphics[width=0.97\textwidth]{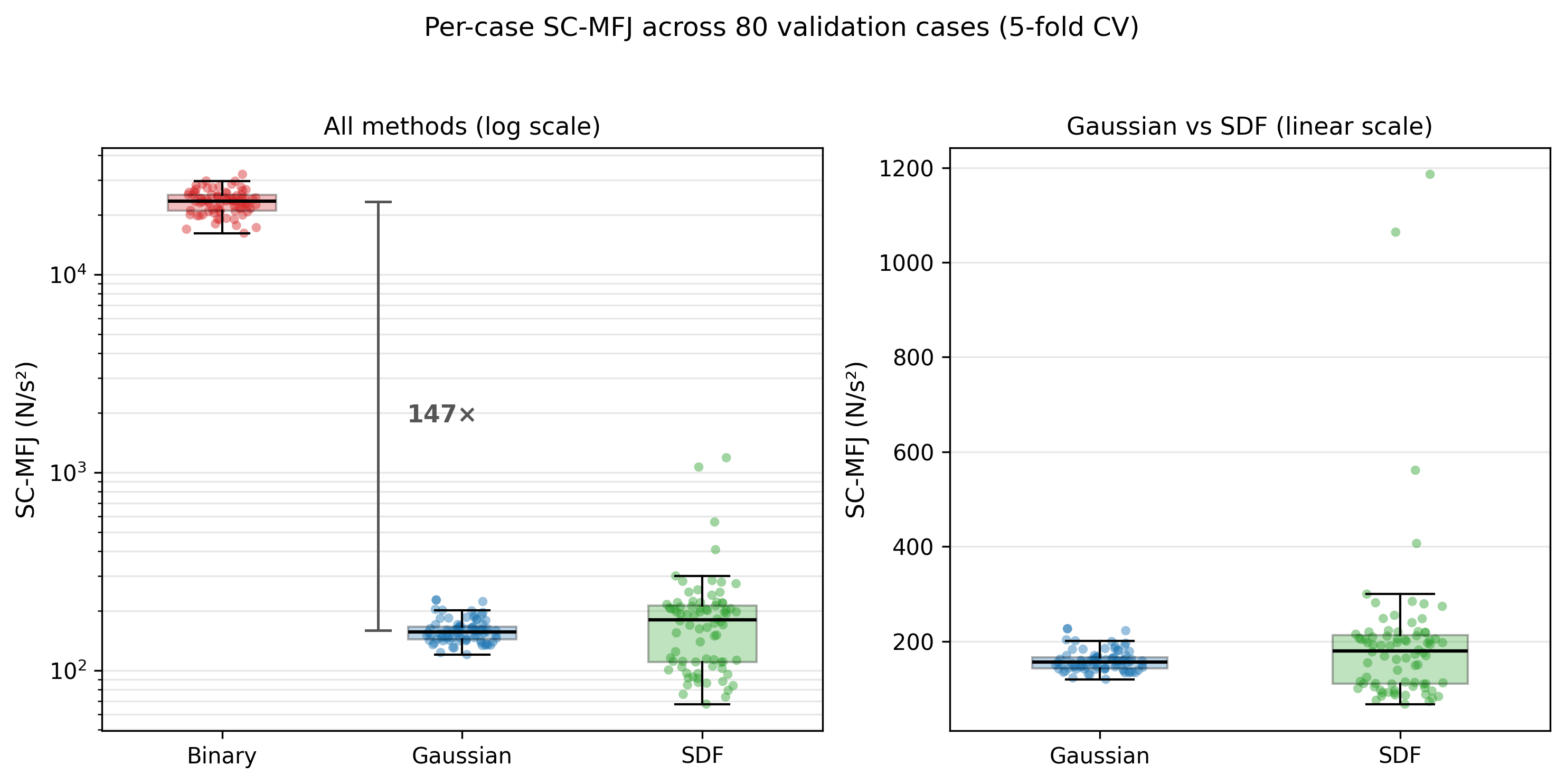}
  \caption{Per-case SC-MFJ across 80 validation cases (five-fold cross-validation).
  Left: all three methods on a log scale; the $147\times$ gap between binary and the smooth methods is visible at a glance.
  Right: Gaussian vs.\ SDF on a linear scale.
  Gaussian smoothing produces consistently low SC-MFJ with minimal outliers; SDF regression shows substantially higher variance, including several cases above 400~N/s$^2$.}
  \label{fig:boxplot}
\end{figure*}
 
\subsection{Visual Example}

Figure~\ref{fig:example} shows a representative case
(PANCREAS\_0021) where Gaussian and SDF achieve similar Dice
scores, isolating the surface quality difference visible to
SC-MFJ.
The binary baseline surface (a) shows clear staircase artifacts.
SDF regression (b) produces a smooth surface.
Gaussian smoothing (c), highlighted in the bottom row, is
visually comparable to SDF---but the force jerk trace (d)
reveals that the binary baseline produces force oscillations
two orders of magnitude larger than either smooth method.

\subsection{Per-Case Distribution}

Figure~\ref{fig:boxplot} shows the per-case SC-MFJ distribution across all 80 validation cases.
The left panel (log scale) makes the $147\times$ gap between the binary baseline and the two smooth methods immediately visible: binary SC-MFJ values cluster above $10^4$~N/s$^2$, while Gaussian and SDF both lie near $10^2$~N/s$^2$.
The right panel (linear scale) zooms in on Gaussian vs.\ SDF.
Gaussian smoothing produces a tight cluster with few outliers; SDF regression produces a wider spread and several high-jerk cases exceeding 400~N/s$^2$, consistent with the fold-level variance reported in Table~\ref{tab:folds}.

\subsection{Parameter Sensitivity}
\label{sec:sensitivity}

Because stiffness $k$ and penetration depth $\delta$ enter SC-MFJ as a linear factor (Eq.~\ref{eq:force}), absolute scores scale with the product $k \cdot \delta$ but relative method rankings should be invariant.
Table~\ref{tab:sensitivity} confirms this empirically.
We evaluated SC-MFJ on Fold~0 (16 cases, 50 trajectories, seed = 42) across nine combinations of $k \in \{0.5, 1.0, 2.0\}$~N/mm and $\delta \in \{0.25, 0.5, 1.0\}$~mm.
The method ordering (Gaussian $<$ SDF $<$ Binary) and the ratio between Binary and Gaussian ($149\times$ on this fold) are preserved exactly across all parameter combinations.
This confirms that SC-MFJ comparisons depend on surface geometry, not on the choice of these contact parameters.

\begin{table}[htb]
\centering
\caption{SC-MFJ (N/s$^2$) under varying stiffness $k$ and penetration depth $\delta$.
Evaluated on Fold~0 (16 cases, 50 trajectories, seed = 42).
Method ranking is preserved across all parameter combinations; scores scale linearly with $k \cdot \delta$.
Within each parameter setting, lower standard deviation indicates
greater case-level consistency; bold marks the more consistent smooth
method.}
\label{tab:sensitivity}
\vspace{2mm}
\small
\resizebox{\columnwidth}{!}{%
\begin{tabular}{@{}cccccc@{}}
\toprule
$k$ (N/mm) & $\delta$ (mm) & Binary & Gaussian & SDF & Ratio B/G \\
\midrule
0.5 & 0.25 & $5{,}971 \pm 764$    & $40 \pm \textbf{7}$    & $41 \pm 58$    & $149\times$ \\
0.5 & 0.50 & $11{,}942 \pm 1{,}527$ & $80 \pm \textbf{13}$   & $82 \pm 117$   & $149\times$ \\
0.5 & 1.00 & $23{,}885 \pm 3{,}054$ & $160 \pm \textbf{27}$  & $163 \pm 233$  & $149\times$ \\
1.0 & 0.25 & $11{,}942 \pm 1{,}527$ & $80 \pm \textbf{13}$   & $82 \pm 117$   & $149\times$ \\
1.0 & 0.50 & $23{,}885 \pm 3{,}054$ & $160 \pm \textbf{27}$  & $163 \pm 233$  & $149\times$ \\
1.0 & 1.00 & $47{,}770 \pm 6{,}109$ & $320 \pm \textbf{53}$  & $326 \pm 466$  & $149\times$ \\
2.0 & 0.25 & $23{,}885 \pm 3{,}054$ & $160 \pm \textbf{27}$  & $163 \pm 233$  & $149\times$ \\
2.0 & 0.50 & $47{,}770 \pm 6{,}109$ & $320 \pm \textbf{53}$  & $326 \pm 466$  & $149\times$ \\
2.0 & 1.00 & $95{,}540 \pm 12{,}217$ & $640 \pm \textbf{106}$ & $653 \pm 933$ & $149\times$ \\
\bottomrule
\end{tabular}}
\end{table}
 
\subsection{The Dice Trade-Off}
 
Gaussian smoothing reduces Dice by 2.6 percentage points relative
to the binary baseline (0.838 vs.\ 0.864).
This is a real cost.
Whether it is acceptable depends on the application: for haptic
simulation, where force discontinuities can degrade force-feedback
stability at 1~kHz, a $147\times$ improvement in force smoothness
may justify the geometric accuracy trade-off.
SC-MFJ does not answer this question---it only makes the trade-off
visible.

\subsection{Gaussian $\sigma$ Sweep}
\label{sec:sigma}

A natural concern is whether the choice of $\sigma = 1.0$ is robust.
To address this, we swept
$\sigma \in \{0.5, 0.75, 1.0, 1.5, 2.0, 3.0\}$ across all 80
cases (Table~\ref{tab:sigma}).
SC-MFJ decreases monotonically with increasing $\sigma$, from
$754 \pm 112$~N/s$^2$ at $\sigma = 0.5$ to $54 \pm 18$~N/s$^2$ at
$\sigma = 3.0$---a $13.9\times$ reduction.
Dice drops more modestly, from 0.839 to 0.799 (4.8\% relative).
This is an expected consequence of isosurface shrinkage under
Gaussian convolution: as $\sigma$ increases, high-frequency boundary
details are smoothed away, reducing both force jerk and spatial
overlap.

\begin{figure}[!t]
  \centering
  \includegraphics[width=0.95\columnwidth]{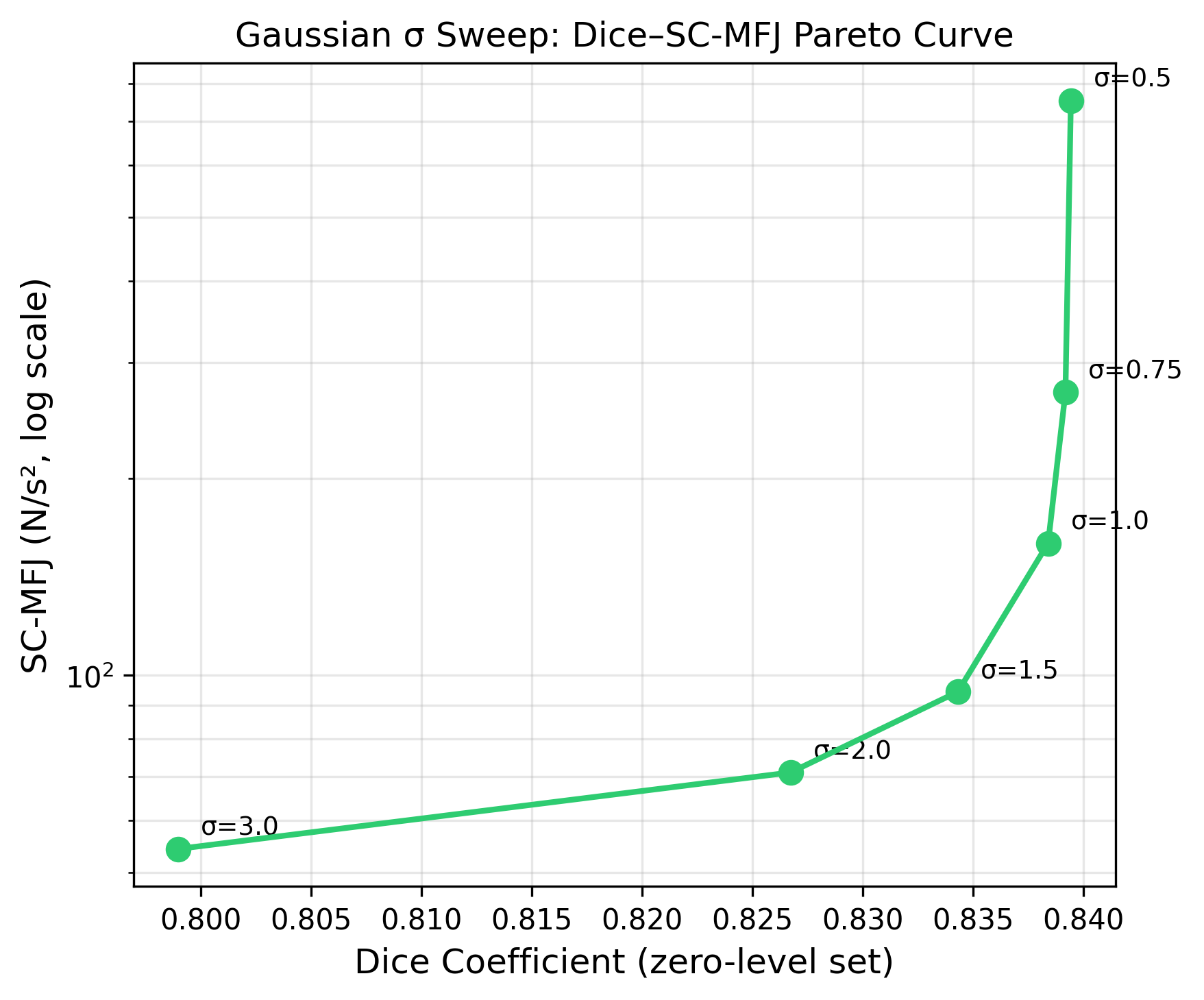}
  \caption{Pareto curve of mean Dice vs.\ mean SC-MFJ across Gaussian
  $\sigma$ values.
  The $\sigma = 1.0$ operating point removes most SC-MFJ while
  preserving Dice; stronger smoothing gives smaller additional gains
  at increasing geometric cost.}
  \label{fig:pareto}
\end{figure}

Figure~\ref{fig:pareto} shows the resulting Pareto curve.
The useful operating region is near $\sigma = 1.0$:
moving from $\sigma = 0.5$ to $\sigma = 1.0$ reduces SC-MFJ
from $754$ to $159$~N/s$^2$ with essentially unchanged Dice,
whereas stronger smoothing gives smaller additional SC-MFJ reductions
while costing more Dice.
We therefore use $\sigma = 1.0$ as a practical operating point, not
as a mathematically unique optimum.

\begin{table}[htb]
\centering
\caption{Gaussian $\sigma$ sweep across 80 cases.
SC-MFJ decreases monotonically with $\sigma$; Dice decreases more
slowly.
All values are mean $\pm$ std.}
\label{tab:sigma}
\vspace{2mm}
\small
\begin{tabular}{@{}ccc@{}}
\toprule
$\sigma$ & Dice & SC-MFJ (N/s$^2$) \\
\midrule
0.50 & $0.839 \pm 0.039$ & $754 \pm 112$ \\
0.75 & $0.839 \pm 0.039$ & $270 \pm 36$ \\
1.00 & $0.838 \pm 0.039$ & $159 \pm 22$ \\
1.50 & $0.834 \pm 0.039$ & $94 \pm 17$ \\
2.00 & $0.827 \pm 0.040$ & $71 \pm 13$ \\
3.00 & $0.799 \pm 0.045$ & $54 \pm 18$ \\
\bottomrule
\end{tabular}
\end{table}

\subsection{Dice--SC-MFJ Correlation}
\label{sec:correlation}

\begin{figure}[!t]
  \centering
  \includegraphics[width=0.95\columnwidth]{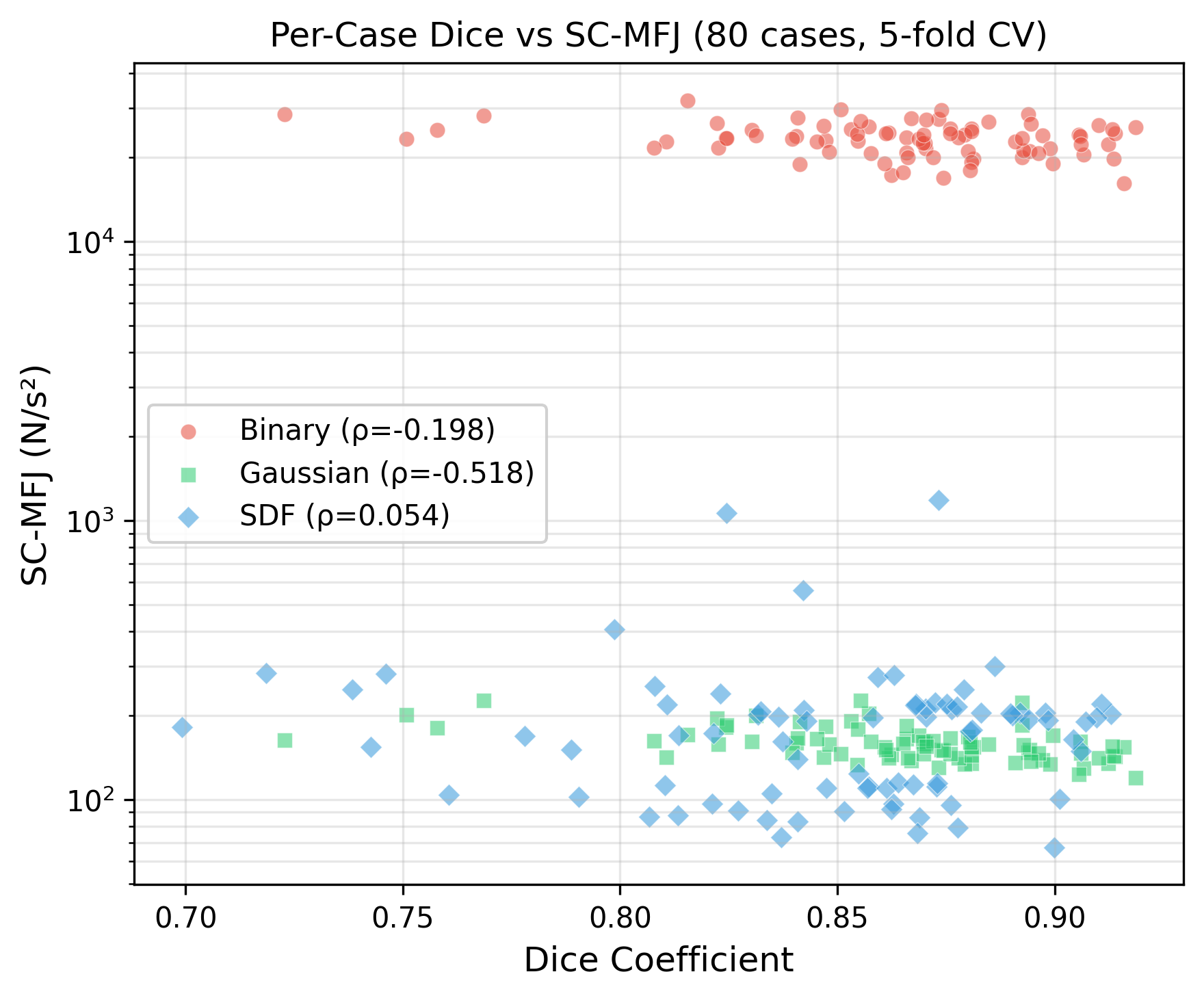}
  \caption{Per-case Dice vs.\ SC-MFJ for 80 cases across three
  methods (log scale for SC-MFJ).
  Spearman $\rho$ values are reported per method.
  The weak-to-moderate correlations---especially $\rho = 0.054$ for SDF---support
  that Dice and SC-MFJ measure largely orthogonal aspects of segmentation
  quality.}
  \label{fig:scatter}
\end{figure}

If SC-MFJ were merely a proxy for Dice, it would add no information.
Figure~\ref{fig:scatter} shows per-case Dice versus SC-MFJ for
all 80 cases across the three methods.
The Spearman rank correlations are weak to moderate: $\rho = -0.198$ for
Binary, $\rho = 0.054$ for SDF regression, and $\rho = -0.518$
for Gaussian.
Even the strongest correlation (Gaussian) leaves most of the rank
variation unexplained.
The near-zero correlation for SDF ($\rho = 0.054$) is particularly
telling: a case's Dice score carries essentially no information about
its haptic quality under SDF regression.
These results confirm that SC-MFJ captures a complementary dimension
of segmentation quality that Dice does not measure.

\subsection{Trajectory Convergence}
\label{sec:convergence}

\begin{figure}[!t]
  \centering
  \includegraphics[width=0.95\columnwidth]{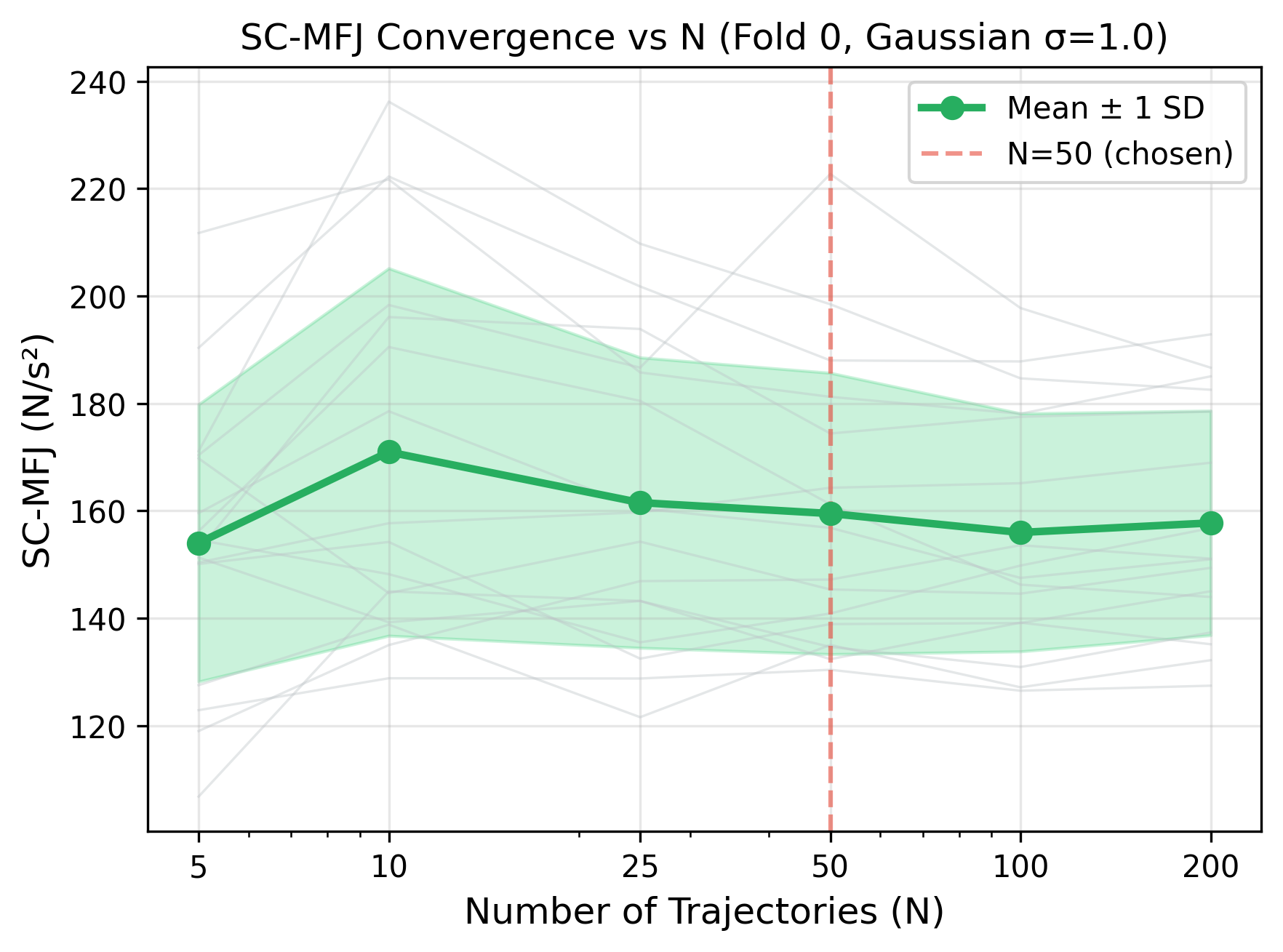}
  \caption{SC-MFJ convergence as a function of the number of
  trajectories $N$ (Fold~0, 16 cases, Gaussian $\sigma = 1.0$).
  The aggregate mean (green line) stabilizes by $N = 50$ (dashed red);
  the shaded band shows $\pm 1$ standard deviation across cases.
  Grey lines show individual cases.}
  \label{fig:convergence}
\end{figure}

SC-MFJ averages over $N$ random surface walks.
To justify the choice of $N = 50$, we evaluated
$N \in \{5, 10, 25, 50, 100, 200\}$ on the 16 cases of Fold~0
using Gaussian $\sigma = 1.0$ (Figure~\ref{fig:convergence}).
The aggregate mean SC-MFJ stabilizes by $N = 50$: the deviation from
the $N = 200$ reference value is $1.1\%$, and the coefficient of
variation decreases from $16.8\%$ at $N = 5$ to $16.4\%$ at
$N = 50$.
Doubling to $N = 100$ or quadrupling to $N = 200$ yields negligible
change in the mean ($< 1.2\%$) while reducing the coefficient of
variation only modestly (to $14.2\%$ and $13.3\%$, respectively).
$N = 50$ thus provides a practical balance between aggregate
measurement stability and computational cost; a full estimator-variance
analysis over repeated random seeds remains future work.

\subsection{Cross-Anatomy Validation: LiTS Liver}
\label{sec:lits}

To assess whether the pattern observed on pancreas generalises to a
different organ, we evaluated SC-MFJ on the LiTS liver
dataset~\cite{Bilic2023} (131 cases, five-fold cross-validation).
Table~\ref{tab:lits} shows the results for the binary baseline and
Gaussian $\sigma = 1.0$.

The binary-to-Gaussian SC-MFJ gap is $189\times$ on liver
($22{,}531 \pm 6{,}086$ vs.\ $119 \pm 76$~N/s$^2$), exceeding the
$147\times$ gap observed on pancreas.
Mean binary Dice is $0.963 \pm 0.038$, substantially higher than on the
more challenging pancreas anatomy, yet the binary baseline still
produces force jerk above $2 \times 10^4$~N/s$^2$---confirming that
high geometric overlap does not guarantee haptic suitability.
The Dice cost of Gaussian smoothing is only 0.8 percentage points
on liver ($0.963 \to 0.955$), less than one-third of the 2.6~pp
cost on pancreas, while HD95 remains essentially unchanged
($12.2 \to 12.0$~mm).
Gaussian SC-MFJ is stable across folds, ranging from 105 to 137~N/s$^2$.

These results show that the core finding---a simple Gaussian
post-processing step eliminates the vast majority of force
jerk---transfers across anatomies and dataset sizes.

\begin{table}[H]
\centering
\caption{LiTS liver results. Values are mean $\pm$ std.
For SC-MFJ, lower standard deviation indicates greater case-level
consistency; bold marks the more consistent SC-MFJ result.}
\label{tab:lits}
\vspace{2mm}
\footnotesize
\setlength{\tabcolsep}{2.5pt}
\begin{tabular}{@{}lccc@{}}
\toprule
Method & Dice & HD95 & SC-MFJ \\
       &      & (mm) & (N/s$^2$) \\
\midrule
Binary   & $0.963 \pm 0.038$ & $12.2 \pm 27.4$ & $22{,}531 \pm 6{,}086$ \\
Gaussian & $0.955 \pm 0.040$ & $12.0 \pm 26.8$ & $119 \pm \textbf{76}$ \\
\bottomrule
\end{tabular}
\end{table}

\section{Discussion}
\label{sec:discussion}
 
SC-MFJ is a limited tool.
It measures one specific thing---force jerk along simulated surface
walks---and nothing else.
It does not replace Dice, HD95, or any other geometric metric;
it complements them for applications where physical force rendering matters.
 
SDF regression has been proposed as an approach for producing smoother organ boundaries~\cite{Xue2020}, making it a natural candidate for haptic rendering applications.
SC-MFJ qualifies this assumption for force-rendering applications: while SDF regression achieves
comparable \emph{mean} haptic quality to Gaussian smoothing, it shows
much higher variability than Gaussian smoothing, both at the case level
($\pm 168$~N/s$^2$ vs.\ $\pm 22$~N/s$^2$) and across folds
(range $194$~N/s$^2$ vs.\ $11$~N/s$^2$).
This means that a practitioner cannot reliably predict haptic quality
from a given trained model.
Gaussian smoothing, using only a single post-processing step, produces
consistently low force jerk across every fold.
 
Several limitations should be noted.
First, the SDF regression analysis is conducted on a single
dataset (NIH Pancreas CT) and a single organ.
The binary-vs-Gaussian comparison, however, is replicated on the LiTS
liver dataset (Section~\ref{sec:lits}), where the $189\times$ gap
confirms that the core finding generalises across anatomies.
The metric itself is organ-agnostic---it operates on any
isosurface---and its formal properties (Section~\ref{sec:formal})
hold independently of anatomy.
The $\sigma$ sweep (Section~\ref{sec:sigma}) further demonstrates
that the metric's behaviour is systematic, not artifact-specific:
SC-MFJ decreases monotonically with smoothing strength across all
80 cases, following a Pareto trade-off with Dice.
Nevertheless, extending the SDF comparison and the $\sigma$ sweep
to additional anatomies remains future work.
Second, we have not validated SC-MFJ against human haptic perception.
A study correlating SC-MFJ scores with expert ratings of haptic
realism would strengthen the metric's practical relevance; this is
planned as future work.
Third, the force model is deliberately simple (constant-depth spring
penalty).
A more sophisticated contact model might capture additional aspects
of haptic quality, but would also introduce more free parameters.
We chose simplicity.

SC-MFJ uses random surface walks rather than task-specific instrument trajectories.
This is deliberate.
The metric is not trying to reproduce one EUS maneuver, needle path,
or surgical tool motion.
It asks a simpler first question: if a haptic tool touches this
segmented surface at many locations, does the surface produce smooth
contact forces or does it produce sharp force oscillations?
Each walk starts from a randomly selected surface vertex and then
moves tangentially while being projected back onto the isosurface.
This spreads the samples over different parts of the organ and avoids
basing the metric on one manually chosen path.
A task-specific trajectory study would answer a different question:
how smooth is force feedback along one particular procedural motion?
That is useful for selected medical applications, but it is separate
from measuring the basic surface quality of the segmentation.

We keep the simulated update rate fixed at 1~kHz, a common haptic rendering standard~\cite{Rizzi2012}.
At fixed sampled trajectory points, changing $\Delta t$ rescales Eq.~\ref{eq:jerk} by $1/\Delta t^2$; if the update rate changes at fixed stylus speed, $\Delta x$ also changes and the sampled normals may differ.
For example, 2~kHz sampling at the same stylus speed would primarily
produce more samples along nearly the same surface path, and we would
expect the qualitative method ranking to remain similar, but this has
not been tested directly.
Thus Table~\ref{tab:sensitivity} should be interpreted only as evidence for invariance to contact parameters $k$ and $\delta$; update-rate sensitivity remains future work.
 
The correlation analysis (Section~\ref{sec:correlation}) provides
direct evidence that SC-MFJ is not redundant with Dice.
The near-zero Spearman correlation for SDF regression
($\rho = 0.054$) is particularly important: it means that selecting
an SDF model based on high Dice provides no assurance of good
haptic quality.
Even the strongest within-method correlation (Gaussian,
$\rho = -0.518$) leaves the majority of SC-MFJ variance
unexplained by Dice.
Together with the main pancreas and liver comparisons, where the large
SC-MFJ gaps are invisible to HD95, this supports the case that SC-MFJ
is not redundant with either Dice or HD95 when the downstream
application involves physical simulation.

The finding that Gaussian smoothing performs well is not itself
surprising---smoothing reduces roughness, which should reduce jerk.
What SC-MFJ adds is a way to \emph{quantify} the effect and compare
it against alternatives.
Without the metric, the community must rely on visual inspection or
expensive user studies to assess the haptic suitability of fine surface
structure in segmentation-derived 3D models.

\section{Conclusion}
\label{sec:conclusion}
 
We presented SC-MFJ, a simple metric for evaluating the haptic
suitability of segmentation-derived 3D models from medical images.
Applied to pancreas CT segmentation and LiTS liver segmentation, it
reveals large binary-to-Gaussian quality gaps of $147\times$ and
$189\times$, respectively---gaps invisible to standard geometric metrics.
It also shows that learned SDF regression, while achieving comparable
mean haptic quality to Gaussian smoothing, does so with substantially
higher variability across training folds.
A Gaussian $\sigma$ sweep reveals a Pareto trade-off between
geometric accuracy and haptic quality, with $\sigma = 1.0$ as a
practical operating point; per-case correlation analysis confirms that SC-MFJ
captures a complementary quality dimension not predicted by Dice,
particularly under SDF regression ($\rho = 0.054$).
Cross-anatomy validation on the LiTS liver dataset confirms these
findings, with an even larger binary-to-Gaussian gap of $189\times$
across 131 cases.
 
SC-MFJ is computationally cheap and runs on CPU.
It is intended as a diagnostic tool for groups building surgical
simulation pipelines: before running an expensive haptic user study,
they can first check whether the segmentation surface is likely to
produce smooth force feedback at all.

\section*{Acknowledgments}
This work is funded by the Deutsche Forschungsgemeinschaft (DFG),
project number MA~6791/1-1.


\begin{thebibliography}{99}
\label{references}
 
 
\bibitem[Bil23a]{Bilic2023}
Bilic, P., Christ, P., Li, H.B., Vorontsov, E., et al.
The Liver Tumor Segmentation Benchmark (LiTS).
\emph{Medical Image Analysis}, 84, p.102680, 2023.

\bibitem[Cha16a]{Chan2016}
Chan, S., Li, P., Locketz, G., Salisbury, K., and Blevins, N.H.
High-fidelity haptic and visual rendering for patient-specific
simulation of temporal bone surgery.
\emph{Computer Assisted Surgery}, 21(1), pp.85--101, 2016.

\bibitem[Col94a]{Colgate1994}
Colgate, J.E. and Brown, J.M.
Factors affecting the Z-width of a haptic display.
In \emph{Proc.\ IEEE Int.\ Conf.\ Robotics and Automation (ICRA)},
pp.3205--3210, 1994.

\bibitem[Est14a]{Estrada2014}
Estrada, S., O'Malley, M.K., Duran, C., Schulz, D.G., and Bismuth, J.
On the development of objective metrics for surgical skills
evaluation based on tool motion.
In \emph{Proc.\ IEEE Int.\ Conf.\ Systems, Man, and Cybernetics},
pp.3144--3149, 2014.

\bibitem[Fla85a]{Flash1985}
Flash, T. and Hogan, N.
The coordination of arm movements: An experimentally confirmed
mathematical model.
\emph{Journal of Neuroscience}, 5(7), pp.1688--1703, 1985.

\bibitem[For13a]{Fortmeier2013}
Fortmeier, D., Mastmeyer, A., and Handels, H.
Optimized image-based soft tissue deformation algorithms for
visualization of haptic needle insertion.
\emph{Studies in Health Technology and Informatics}, 184,
pp.136--140, 2013.

\bibitem[Ise21a]{Isensee2021}
Isensee, F., Jaeger, P.F., Kohl, S.A.A., Petersen, J., and
Maier-Hein, K.H.
nnU-Net: a self-configuring method for deep learning-based
biomedical image segmentation.
\emph{Nature Methods}, 18(2), pp.203--211, 2021.

\bibitem[Lin08a]{Lin2008}
Lin, M.C. and Otaduy, M.A.
\emph{Haptic Rendering: Foundations, Algorithms, and Applications}.
A K Peters, 2008.

\bibitem[Lor87a]{Lorensen1987}
Lorensen, W.E. and Cline, H.E.
Marching Cubes: A high resolution 3D surface construction algorithm.
\emph{ACM SIGGRAPH Computer Graphics}, 21(4), pp.163--169, 1987.

\bibitem[Ma20a]{Ma2020}
Ma, J., Wei, Z., Zhang, Y., Wang, Y., et al.
How distance transform maps boost segmentation CNNs: An empirical
study.
In \emph{Proc.\ Medical Imaging with Deep Learning (MIDL)},
pp.479--492, 2020.

\bibitem[Mai24a]{MaierHein2024}
Maier-Hein, L., Reinke, A., Godau, P., Tizabi, M.D., et al.
Metrics reloaded: recommendations for image analysis validation.
\emph{Nature Methods}, 21(2), pp.195--212, 2024.

\bibitem[Mas12a]{Mastmeyer2012}
Mastmeyer, A., Fortmeier, D., and Handels, H.
Anisotropic diffusion for direct haptic volume rendering in lumbar
puncture simulation.
In \emph{Proc.\ Bildverarbeitung f\"{u}r die Medizin (BVM)},
pp.286--291, 2012.

\bibitem[Mas13a]{Mastmeyer2013}
Mastmeyer, A., Hecht, T., Fortmeier, D., and Handels, H.
Ray-casting-based evaluation framework for needle insertion force
feedback algorithms.
In \emph{Proc.\ Bildverarbeitung f\"{u}r die Medizin (BVM)},
pp.3--8, 2013.

\bibitem[Nik21a]{Nikolov2021}
Nikolov, S., Blackwell, S., Zverovitch, A., Mendes, R., et al.
Clinically applicable segmentation of head and neck anatomy for
radiotherapy: deep learning algorithm development and validation
study.
\emph{Journal of Medical Internet Research}, 23(7), p.e26151, 2021.

\bibitem[Nys17a]{Nysjo2017}
Nysj\"{o}, F., Olsson, P., Malmberg, F., Carlbom, I.B., and
Nystr\"{o}m, I.
Using anti-aliased signed distance fields for generating surgical
guides and plates from CT images.
\emph{Journal of WSCG}, 25(1), pp.11--20, 2017.

\bibitem[Riz12a]{Rizzi2012}
Rizzi, S.H., Luciano, C.J., and Banerjee, P.
Comparison of algorithms for haptic interaction with isosurfaces
extracted from volumetric datasets.
\emph{ASME J.\ Comput.\ Inf.\ Sci.\ Eng.}, 12(2), p.021004, 2012.

\bibitem[Rot15a]{Roth2015}
Roth, H.R., Lu, L., Farag, A., Shin, H.-C., Liu, J., Turkbey, E.B.,
and Summers, R.M.
DeepOrgan: Multi-level deep convolutional networks for automated
pancreas segmentation.
In \emph{Proc.\ MICCAI}, pp.556--564, 2015.

\bibitem[Sha23a]{Shayan2023}
Shayan, A.M., Singh, S., Gao, J., et al.
Measuring hand movement for suturing skill assessment:
A simulation-based study.
\emph{Surgery}, 174(5), pp.1184--1192, 2023.

\bibitem[Shi21a]{Shit2021}
Shit, S., Paetzold, J.C., Sekuboyina, A., et al.
clDice---A novel topology-preserving loss function for tubular
structure segmentation.
In \emph{Proc.\ IEEE/CVF Conf.\ Computer Vision and Pattern
Recognition (CVPR)}, pp.16560--16569, 2021.

\bibitem[Xue20a]{Xue2020}
Xue, Y., Tang, H., Qiao, Z., et al.
Shape-aware organ segmentation by predicting signed distance maps.
In \emph{Proc.\ AAAI Conference on Artificial Intelligence},
pp.12565--12572, 2020.

\bibitem[Yan23a]{Yang2023}
Yang, H., Sun, Y., Sundaramoorthi, G., and Yezzi, A.
StEik: Stabilizing the optimization of neural signed distance
functions and finer shape representation.
In \emph{Advances in Neural Information Processing Systems
(NeurIPS)}, pp.13993--14004, 2023.
\end{thebibliography}
\end{document}